\documentclass[sn-mathphys,Numbered]{sn-jnl}

\usepackage{graphicx}%
\usepackage{multirow}%
\usepackage{amsmath,amssymb,amsfonts}%
\usepackage{amsthm}%
\usepackage{mathrsfs}%
\usepackage[title]{appendix}%
\usepackage{xcolor}%
\usepackage{textcomp}%
\usepackage{manyfoot}%
\usepackage{booktabs}%
\usepackage{algorithm}%
\usepackage{algorithmicx}%
\usepackage{algpseudocode}%
\usepackage{listings}%
\usepackage{longtable}%

\theoremstyle{thmstyleone}%
\theoremstyle{thmstyletwo}%

\theoremstyle{thmstylethree}%

\begin{document}

\title[Article Title]{AI-Driven Approaches for Optimizing Power Consumption: A Comprehensive Survey}

\author[1]{\fnm{Parag} \sur{Biswas}}\email{text2parag@gmail.com}

\author[2]{\fnm{Abdur} \sur{Rashid}}\email{rabdurrashid091@gmail.com}

\author[3]{\fnm{Angona} \sur{Biswas}}\email{angonabiswas28@gmail.com}

\author*[4]{\fnm{Md Abdullah Al} \sur{Nasim}}\email{nasim.abdullah@ieee.org}

\author[5]{\fnm{Kishor} \sur{Datta Gupta}}\email{kgupta@cau.edu}

\author[6]{\fnm{Roy} \sur{George}}\email{george@cau.edu}

\affil[1, 2]{\orgdiv{MSEM Department}, \orgname{Westcliff university},  \orgaddress{{\city{California}, \country{United States}}}}

\affil[3, 4]{\orgdiv{Research and Development Department}, \orgname{Pioneer Alpha},  \orgaddress{{\city{Dhaka}, \country{Bangladesh}}}}

\affil[5, 6]{\orgdiv{Department of Computer and Information Science}, \orgname{Clark Atlanta University}, {\city{Georgia}, \country{USA}}}

\abstract{Reduced environmental effect, lower operating costs, and a stable and sustainable energy supply for current and future generations are the main reasons why power optimization is important. Power optimization makes ensuring that energy is used more effectively, cutting down on waste and optimizing the utilization of resources.In today's world, power optimization and artificial intelligence (AI) integration are essential to changing the way energy is produced, used, and distributed. Real-time monitoring and analysis of power usage trends is made possible by AI-driven algorithms and predictive analytics, which enable dynamic modifications to effectively satisfy demand. Efficiency and sustainability are increased when power consumption is optimized in different sectors thanks to the use of intelligent systems. This survey paper comprises an extensive review of the several AI techniques used for power optimization as well as a methodical analysis of the literature for the study of various intelligent system application domains across different disciplines of power consumption.This literature review identifies the performance and outcomes of 17 different research methods by assessing them, and it aims to distill valuable insights into their strengths and limitations. Furthermore, this article outlines future directions in the integration of AI for power consumption optimization.}




\maketitle

\section{Introduction}\label{sec1}
One of the most difficult problems of the twenty-first century is undoubtedly the reliance on fossil fuels as the main source of energy and the accompanying environmental difficulties (such as climate change). As a result, developing solutions to address global energy challenges is urgently needed in a variety of areas \cite{ngarambe2020use}. The amount of electrical energy used by systems, industries, or devices over a given time period is referred to as power consumption. This measure, which includes the energy needed to drive industrial machines, run electronic gadgets, and power important services, is a crucial component of modern life. Power consumption has historically developed in tandem with the industrial revolution, which was typified by the widespread use of electrical systems in the late 19th and early 20th centuries \cite{sorrell2015reducing}. The growing reliance on electricity for production, communication, and illumination signified a dramatic change in the energy requirements of society. Data from the past shows a notable increase in power use over the years. Energy consumption was relatively minimal in the early stages of electrification in the late 1800s, mostly restricted to lights and a few industrial operations. On the other side, power optimization refers to the intentional management and reduction of power usage without sacrificing functionality or performance. The idea of power optimization became more well-known as worries about environmental sustainability and energy conservation increased. Evidence from the past shows a slow transition towards power optimization techniques, especially with the introduction of energy-saving devices and methods. The necessity for sustainable energy consumption methods was brought to light by the Kyoto Protocol in 1997 and other international initiatives \cite{economidou2020review}, which prompted businesses and governments to look into creative ways to optimize power usage.

\begin{figure}[H]
\includegraphics[width= 11.5 cm]{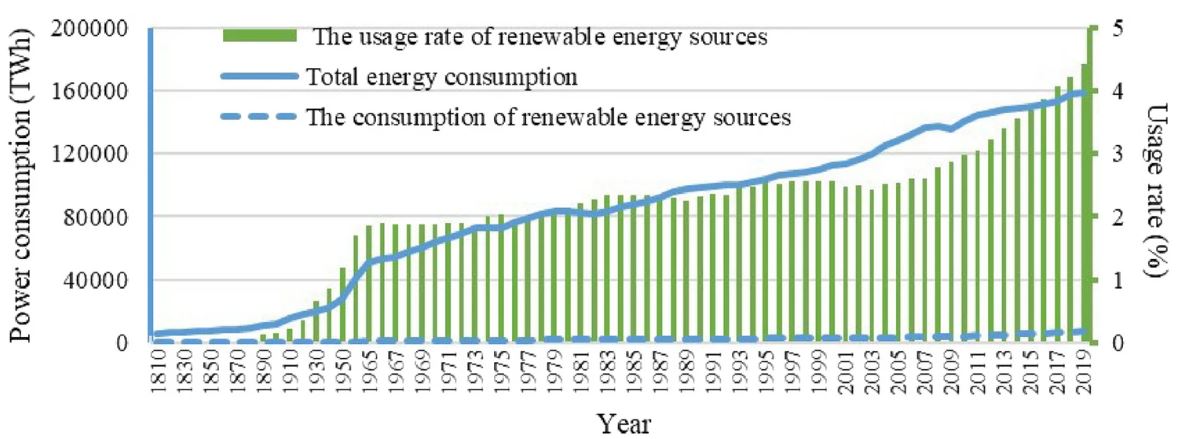}
\caption{Power consumption worldwide and use of renewable energy. \cite{wang2023ai}
\label{fig1}}
\end{figure}  
\unskip

\begin{figure}[H]
\includegraphics[width= 11.5 cm]{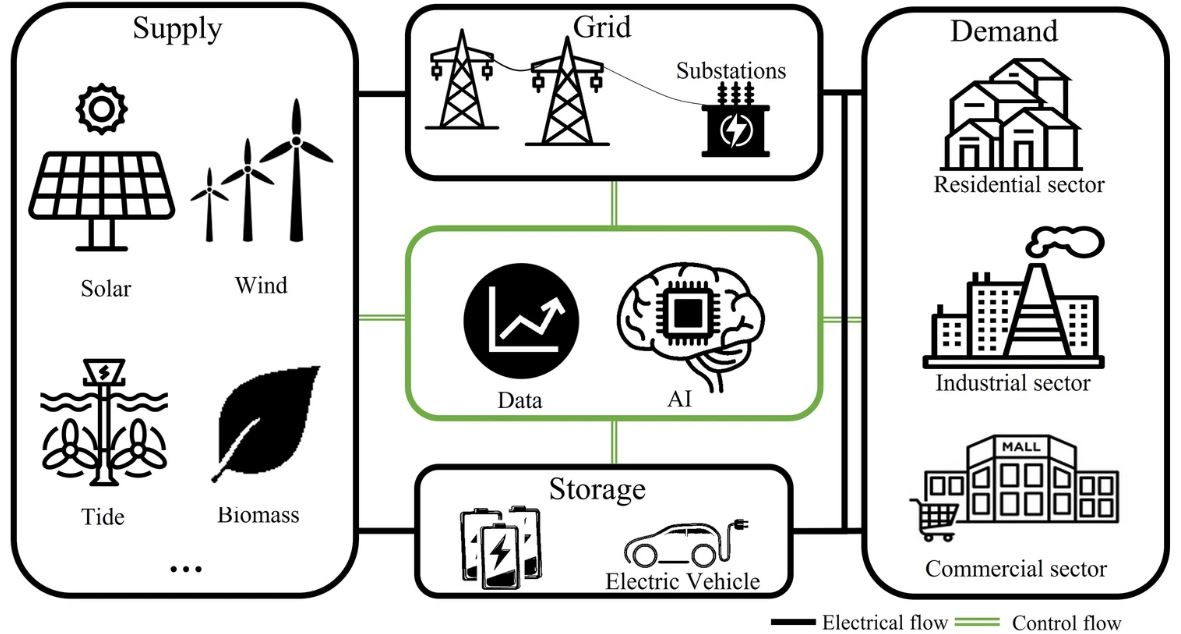}
\caption{Sustainable energy supply, intelligent energy use, sophisticated grid analytics, mobile and stationary energy storage, and real-time control and management with AI and data at the core are all features of a typical smart grid architecture. \cite{wang2023ai}
\label{fig2}}
\end{figure}

The past record of global power consumption data from 1800 to 2019 is depicted in Figure \ref{fig1}, which demonstrates the rapid increase in the share of energy provided by RESs, notwithstanding their relative smallness. Together with nuclear power, renewable energy sources (RESs) will, on average, satisfy more than 90\% of the growth in worldwide demand by 2025, according to the International Energy Agency's Electricity Market Report 2023 \cite{wang2023ai}. The current era's extensive use of smart grids, energy-efficient appliances, and green construction practices are indications of power optimization initiatives.  Energy monitoring systems could undergo many revolutions thanks to artificial intelligence (AI) \cite{mischos2023intelligent}. Energy monitoring systems are essential for effectively tracking and controlling energy use, cutting expenses, and limiting environmental effects. These are a few of the major roles AI plays towards this change \cite{mischos2023intelligent}. A smart grid that uses AI can balance the production and consumption of electricity, maximize the use of renewable resources, increase grid dependability, and guarantee security. The market share of smart grid applications has increased significantly in the last few years. Figure \ref{fig2} shows the general layout of a smart grid. 

\unskip

\begin{figure}[H]
\includegraphics[width= 11.5 cm]{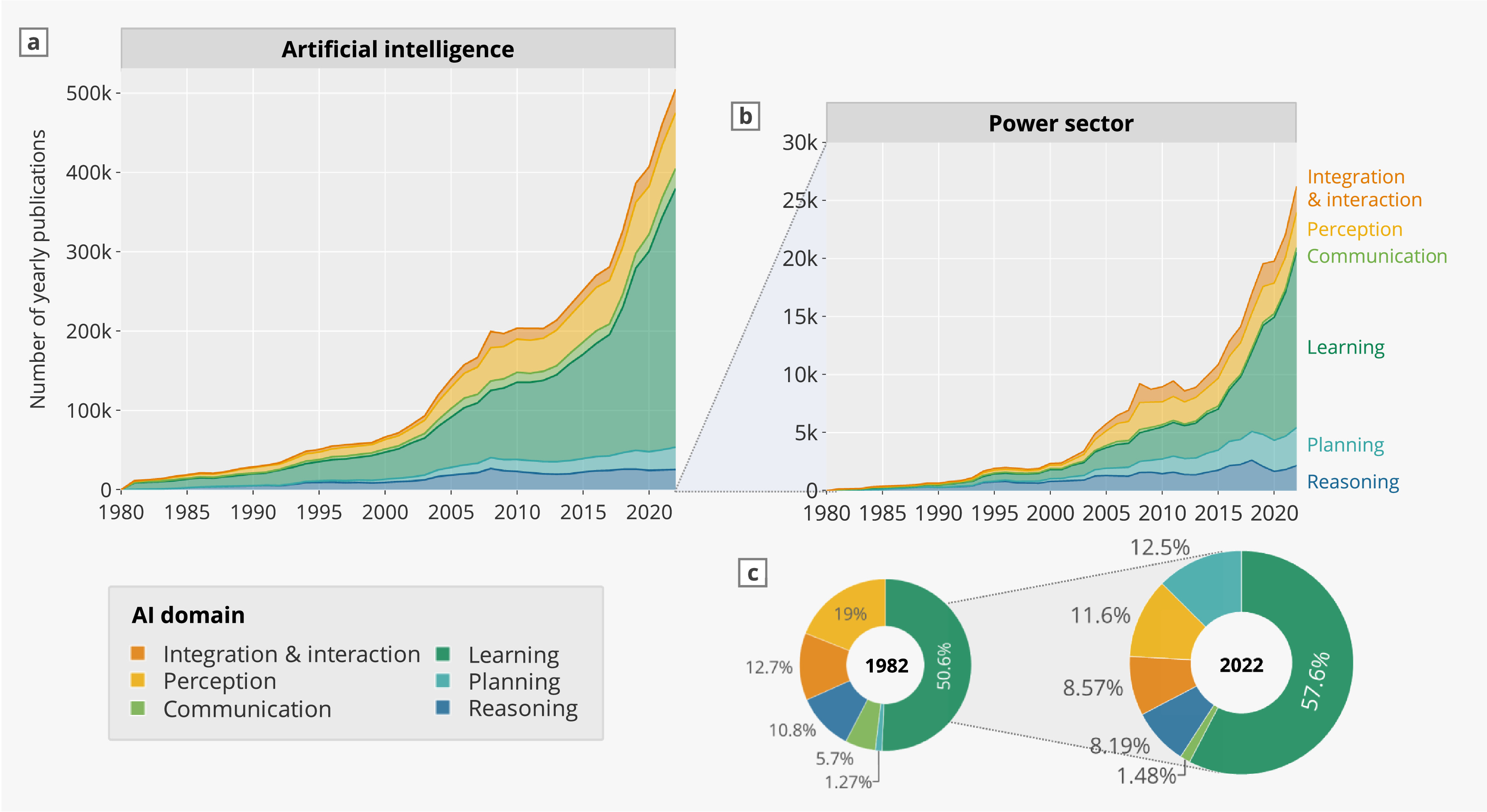}
\caption{Sustainable energy supply, intelligent energy use, sophisticated grid analytics, mobile and stationary energy storage, and real-time control and management with AI and data at the core are all features of a typical smart grid architecture. \cite{heymann2024reviewing}
\label{result}}
\end{figure}  
\unskip

To show the evolution of AI use cases in power systems is elaborated by the research \cite{heymann2024reviewing}. The authors' study of 258'919 papers published between 1982 and 2022 is based on online searches.
 They took into account six AI areas and 19 use cases in the supply chain for power systems. where retail, generation, and transmission are the power system supply chain segments that have been studied the most. According to their findings, the AI domains with the greatest study coverage are "planning" (14\%) and "learning" (45\%). They also examined how the present taxonomy of AI impedes thorough research on advantages and disadvantages. The overview of their research is shown in Figure \ref{result} where authors explain their overall findings. When examining the distribution of AI applications throughout the various components of the power system supply chain, the industry with the greatest number of yearly publications (54.5\% in 2022) is the retail sector, followed by generation and transmission networks. 

The above evaluation includes a few previous study themes and main driving forces. This survey paper's main motivation and survey structure are listed below.

\subsection{Motivation}
Artificial Intelligence (AI) is revolutionizing power resource management and utilization, hence its incorporation into power consumption studies is extremely important. AI's dynamic optimization skills promote energy efficiency in a variety of fields by enabling real-time power usage modifications based on changing needs and environmental factors. This survey paper places a significant emphasis on exploring diverse sectors where intelligent systems play a crucial part in optimizing power consumption. Furthermore, the purpose is to present an in-depth overview of the AI techniques utilized in previous power consumption research. This research seeks to outline the wide range of AI-driven methods and approaches that have been applied in the context of power optimization by evaluating and synthesizing the available literature. This study serves as a learning resource by providing an overview of the AI techniques utilized in earlier power consumption studies. 

\subsection{Survey Framework}

This study reviews state-of-the-art techniques that apply algorithms based on AI for power optimization across a variety of application areas.
The research listed below was included:
\begin{itemize}
\item [-] research on Solar System Photovoltaics using AI techniques \cite{boubaker2023assessment}, \cite{guo2020fault}, \cite{augbulut2020performance}
\item [-] studies that manage Thermal Energy using AI techniques \cite{kannari2023energy}, \cite{yang2020model}
\item [-] research using AI techniques for Smart Grids \cite{ahmed2020machine}, \cite{karimipour2019deep}
\item [-] research using AI techniques for  Industrial Automation  \cite{al2022machine} 
\item [-] research evaluating power consumption using the Fuzzy Logic control approach \cite{shah2020fuzzy}, \cite{boujoudar2023fuzzy}
\item [-] research that examine power usage using the Reinforcement Learning approach \cite{malta2023using} 
\item [-] studies that evaluate power consumption using Genetic Algorithm \cite{wahid2020energy}, \cite{khan2020genetic} 
\item [-] studies that calculate power consumption using the Swarm Intelligence technique \cite{ali2023power}
\item [-] studies that analyze power usage using the Machine Learning technique  \cite{tekin2023energy}
\item [-] studies that evaluate power usage using the Neural Network technique \cite{ruan2022hybrid}
\item [-] research that measure power usage using the Predictive Analytics technique \cite{lin2022predictive}
\end{itemize}

\begin{figure}[H]
\includegraphics[width= 13 cm]{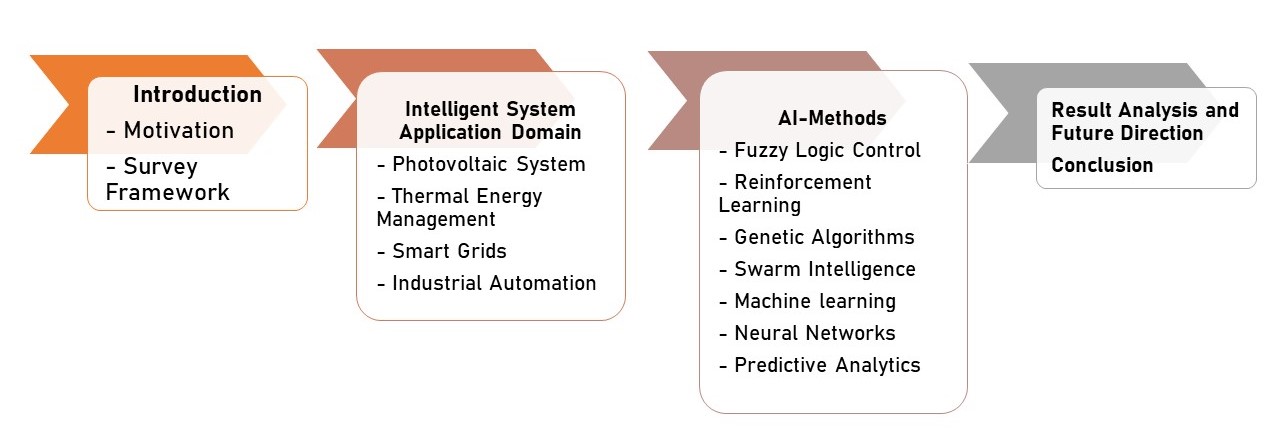}
\caption{Visual depiction of this survey article's structure. 
\label{fig3}}
\end{figure}  
\unskip

Figure \ref{fig3} presents the survey's visual organization. An overview of power consumption and handling procedures from earlier research is provided in Section \ref{sec1}. Additionally, the primary motivation and the paper's structure are shown in this part. The paper's Section \ref{sec2} emphasizes the area of power consumption from earlier studies where intelligent systems are used. Section \ref{sec3} presents an analysis of different approaches used to apply AI concepts in research papers. This survey paper's Section \ref{sec4} will highlight previous findings and outline obstacles that need to be solved moving forward.  Lastly, an essay summarizing the whole survey is included in Section \ref{sec5}. 

\section{Application Domains: Intelligent Systems in Diverse Power Consumption Fields} \label{sec2}

AI technologies that are now being researched and those that are anticipated to emerge soon will be carefully vetted. AI technology is now mostly utilized in the energy sector for diagnosis and prediction. Forecasts pertaining to new energy generation and user-side energy consumption behavior make up the majority of the forecasts. We investigate the numerous application fields of intelligent systems in relation to power consumption in our survey study. Our survey seeks to provide a thorough grasp of the ways in which intelligent systems influence and change the dynamics of power consumption in four variety of disciplines through this investigation.

\subsection{Photovoltaic System}
\begin{figure}[H]
\includegraphics[width= 13 cm]{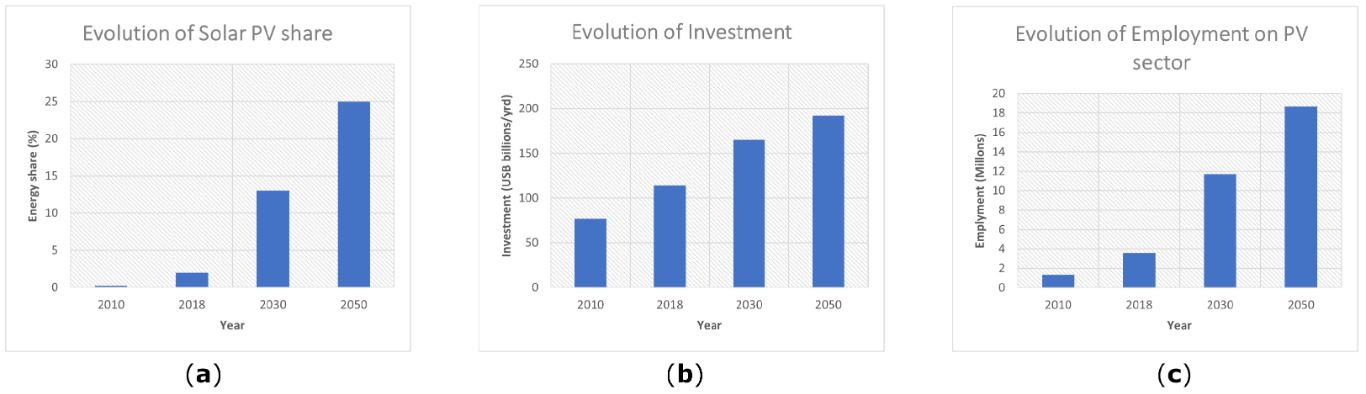}
\caption{The PV sector's significance continues to grow  (a) The evolution of PV systems' energy share; (b) The evolution of PV energy investments; and (c) The evolution of PV industry employment. \cite{mateo2022applications}
\label{fig5}}
\end{figure}

One of the most significant renewable energy sources is solar energy, in which governments and corporations are investing more money each year. Photovoltaic (PV), thermal, and solar energy are the three types of solar energy \cite{mateo2022applications}. The latter has been progressively increasing in recent years and has been shown to be more beneficial and lucrative for industrial production. AI is utilized to address the most significant issues with photovoltaic systems.  Even while AI techniques have several drawbacks, such the volume of data they require and the lengthy calculation periods required for training, they outperform conventional methods in terms of performance \cite{mateo2022applications}. Research is continually being done to identify improved performance strategies and find solutions to these issues. Figure \ref{fig5} illustrates how the percentage of photovoltaic power is rising, and in the next years, it is anticipated to rank among the main energy sources. 

\unskip

In the research study \cite{boubaker2023assessment}, the authors explores fault detection and diagnosis in photovoltaic (PV) modules using machine learning (ML) and deep learning (DL) techniques, specifically focusing on infrared thermography (IRT) images. The investigation encompasses a comprehensive analysis of binary classification (faulty vs. healthy) and multiclass classification (four types of faults). The primary goal of the research is to enhance the safety and efficiency of PV plants by developing robust fault detection and diagnosis models for PV modules. The study aims to contribute to the mitigation of power losses and the maintenance of PV systems by leveraging advanced ML and DL techniques, specifically applied to IRT images. Two methods for using DeepCNN-based processing of IRT datasets for fault identification and classification are shown in the flowchart below (Figure \ref{fig6}) which is proposed by authors: 

\begin{figure}[H]
\includegraphics[width= 11 cm]{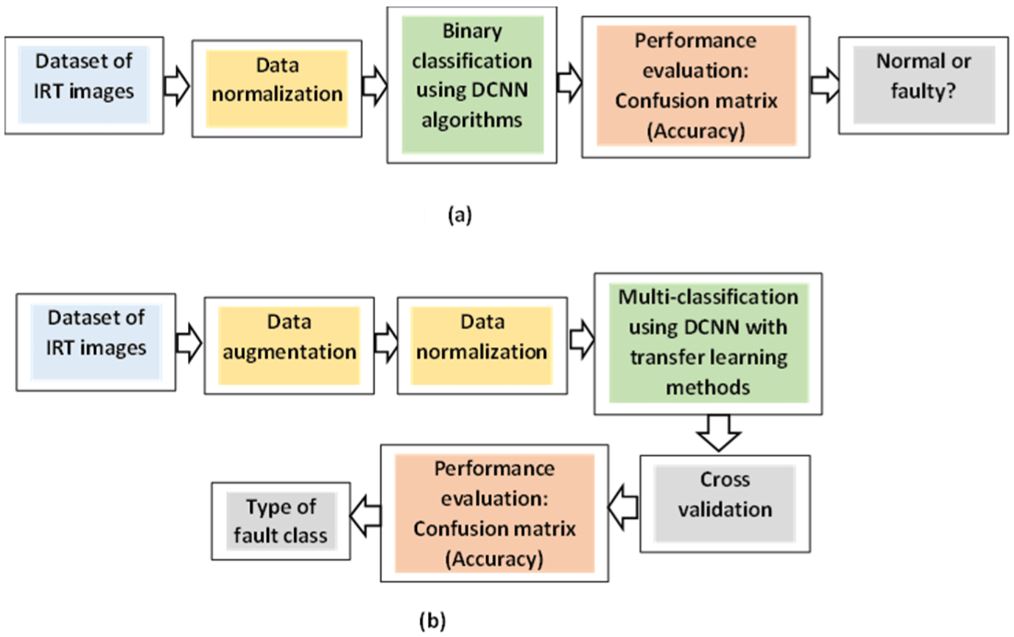}
\caption{The suggested process for employing the IRT dataset for fault classification (b) and detection (a). \cite{boubaker2023assessment}
\label{fig6}}
\end{figure}  
\unskip

The workflow for fault detection and diagnosis of PV modules using infrared thermography (IRT) images involves several key steps. Initially, a dataset containing IRT images of both normal and faulty PV modules is collected. This dataset is then segmented into two sub-datasets: one for binary classification (normal vs. faulty) and another for multiclass classification, categorizing faulty images into four classes representing distinct faults.  Next, two sub-datasets are made: one for multiclass classification (four different fault kinds) and another for binary classification (normal vs. defective).   Models for defect diagnosis and detection are developed using machine learning (ML) and deep learning (DL) approaches, with a Convolutional Neural Network (CNN) configuration receiving special attention. For the experiment, the authors use the KNN, SVM, CatBoost, CNN-SVM, and VGG-16 models. Precision, recall, F1-score, accuracy, and other convolution matrix measures are used to evaluate the models' performance. For both categorization scenarios, a deepCNN setup showed better accuracy, reaching 98.70\%. 

\begin{figure}[H]
\includegraphics[width= 13 cm]{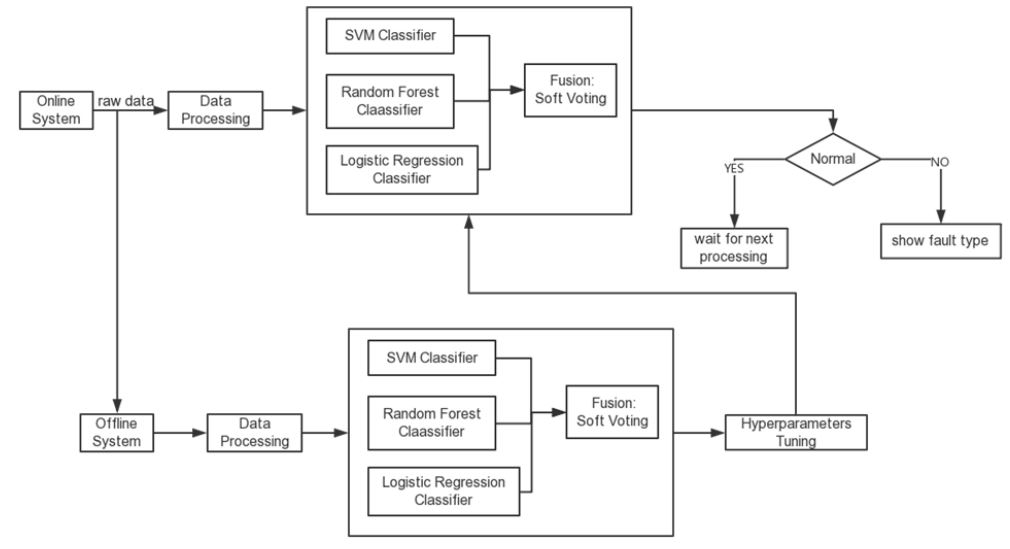}
\caption{Fusion model structure of proposed \cite{guo2020fault} fault diagnosis system. 
\label{fig7}}
\end{figure}

 Machine learning model fusion is used to diagnose fault in photovoltaic system by the authors \cite{guo2020fault}. With the importance of energy concerns only increasing and non-renewable resources running out, this study \cite{guo2020fault} attempts to solve the problems that dispersed solar power plants confront. The research proposes a comprehensive system for real-time fault diagnosis in distributed photovoltaic power stations. The system is designed to process data, provide real-time online diagnoses, retrain models offline at regular intervals, and employ a fusion model of multiple machine learning algorithms for optimal performance. The imported raw data from distributed photovoltaic power stations undergoes a tailored processing pipeline. This includes initial diagnosis, filtering, feature engineering, restructuring, and feature selection. These steps collectively organize and optimize the data, preparing it for subsequent model training. The process of feature engineering involves restructuring the data to retain crucial aspects, such as the average current and voltage of all three phases. Additionally, new features are introduced, encompassing the maximum absolute difference between phases and the median values for current, voltage, power, and energy across all equipment at the same moment. The model training phase adopts a fusion model comprising various classic machine learning classification algorithms. Soft voting, an ensemble technique, is implemented to optimize performance by assigning weights to individual classifiers. This approach allows for a collaborative and robust learning process, harnessing the strengths of different algorithms to create a powerful and versatile predictive model. A framework is shown in Figure \ref{fig7} to further clarify the overall model's guiding concept.

\unskip

 In the model evaluation stage, the performance of distinct models—Support Vector Machine (SVM), Logistic Regressor, Random Forest Regressor, and the proposed fusion model—is scrutinized. The assessment involves analyzing training and cross-validation scores to gauge accuracy and identify potential issues like overfitting. The findings reveal that the fusion model surpasses individual models, achieving notable accuracy scores exceeding 0.95. Notably, SVM demonstrates outstanding performance, especially with approximately 2000 training examples, while Logistic Regressor lags slightly in accuracy. Although the Random Forest Regressor exhibits strong performance, concerns about overfitting are alleviated through the application of the fusion model. The fusion model's stability in cross-validation and its upward trajectory indicate promising results, suggesting further improvements with additional examples.

 The primary objective of  this \cite{augbulut2020performance} study is to enhance the understanding and performance evaluation of concentrated photovoltaic (CPV) systems. The research specifically focuses on two key aspects: the determination of an optimum area ratio for a V-trough concentrator and the prediction of power outputs in CPV systems using machine learning algorithms. The study is conducted in two stages. In the first stage, different-sized layers are designed and mounted on the CPV system to determine the optimum area ratio (AR). This involves gradual mounting of layers to change the concentrator length, impacting the concentration ratio of the system. In the second stage, the power outputs measured in the study are predicted using four machine-learning algorithms: support vector machine (SVM), artificial neural network (ANN), kernel-nearest neighbor (k-NN), and deep learning (DL). Training data include experiment time, concentrated and non-concentrated solar radiation, and PV module temperature. 

The study \cite{augbulut2020performance}  reveals that implementing a double-layer application enhances the concentrator's power output by 16\%, with an average concentration ratio of 1.8. The introduced power coefficient approach proves more dependable than traditional module efficiency data for assessing concentrated photovoltaic (CPV) systems. Determining an optimum area ratio (AR) of 9 for the V-trough concentrator, associated with a calculated power coefficient of 1.35, provides valuable optimization insights. Among machine learning algorithms, Support Vector Machine (SVM) stands out, demonstrating superior prediction accuracy with an R² of 0.9921, low RMSE (0.7082), MBE (0.3357), and MABE (0.6238). These findings collectively enhance our understanding of CPV system performance and offer practical guidance for efficiency optimization.

\subsection{Thermal Energy Management}

The study \cite{kannari2023energy} addresses the pressing concern of the energy crisis, particularly the winter electricity shortage, leading to increased costs for building owners. The primary objective is to leverage near real-time energy performance data to optimize the energy flexibility of buildings. The research focuses on employing a reinforcement learning (RL)-based method for HVAC control, specifically heating, to minimize electricity costs, shift usage away from peak hours, and ensure indoor thermal comfort. The study introduces a deep reinforcement learning-based model for optimizing HVAC systems. The model considers dynamic electricity costs, weather information, and occupant preferences to minimize energy costs while maintaining thermal comfort. Simulations are conducted using electrically heated single-family houses. The RL-based model is tested under varying electricity prices to assess its effectiveness in achieving cost savings without compromising indoor comfort.

\begin{figure}[H]
\includegraphics[width= 13 cm]{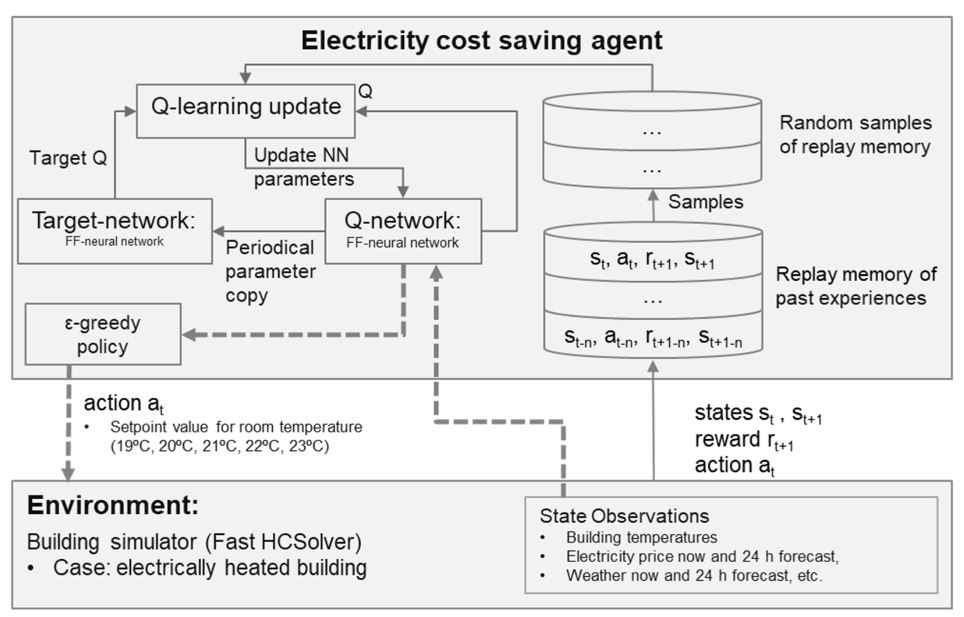}
\caption{The applied reinforcement learning scenario for determining the ideal electric heating setpoint settings for the following day in order to get the best possible reward—a reduction in power bills. \cite{kannari2023energy} 
\label{fig8}}
\end{figure}  
\unskip

Figure \ref{fig8} illustrates the applied reinforcement learning instance for determining the ideal next-hour electric heating threshold values in order to get the optimal reward (savings on power bills). The study reports significant heating cost savings with RL, comparable to reducing the stable indoor temperature setpoint. The algorithm's performance improves over the years, attributed to the agent's learning curve and changes in electricity price dynamics.

The research \cite{yang2020model} aims to develop and evaluate a Model Predictive Control (MPC) system utilizing adaptive machine-learning-based building models for building automation and control applications. The primary goal is to optimize the energy efficiency and indoor thermal comfort of air-conditioning and mechanical ventilation (ACMV) systems in real-world testbeds. The study proposes an innovative approach to building dynamics modeling using a dynamic artificial neural network with a nonlinear autoregressive exogenous structure. This adaptive machine-learning-based model is designed to regularly update itself using online building operation data, offering a dynamic representation of the building's behavior. 

\begin{figure}[H]
\includegraphics[width= 13 cm]{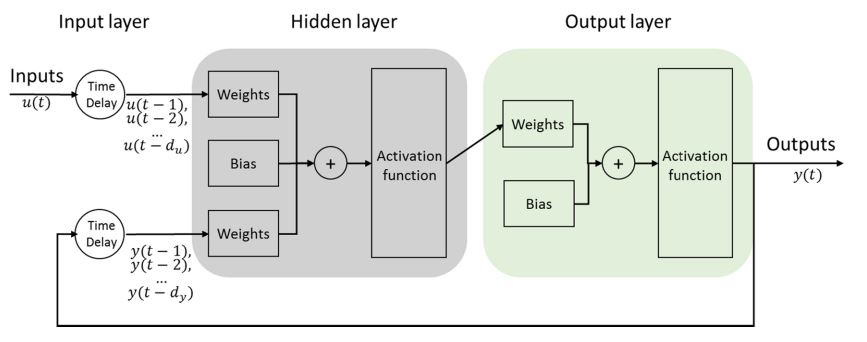}
\caption{A representation of the NARX ANN's schematic design for modeling dynamic systems. \cite{yang2020model}
\label{fig9}}
\end{figure}  
\unskip

This work uses an ANN with the dynamic recurrent nonlinear autoregressive exogenous (NARX) architecture, as seen in Figure \ref{fig9}. The feedforward multi-layer perceptron (MLP) ANN, an ANN structure for modeling continuous linear/nonlinear systems, is the internal architecture of the NARX ANN.  The MPC system demonstrates significant energy efficiency improvements, reducing cooling thermal energy consumption by 58.5\% in the office and cutting cooling electricity consumption by 36.7\% in the lecture theatre. Simultaneously, indoor thermal comfort is notably enhanced.

\subsection{Smart Grids}
The present energy system is heading toward a future where consumers may buy whatever they require, sell it when it becomes too much, and exchange their purchasing rights for other forward-thinking consumers (prosumers) as the smart grid develops \cite{tushar2020challenges}. Global power networks must simultaneously meet the ever-increasing demand for energy while maintaining a steady supply of electricity \cite{strasser2014review}. In smart grids, intelligent systems are used for real-time distribution optimization, control, and monitoring. This covers defect detection, predictive maintenance, and demand response. 

The main objective of the project \cite{ahmed2020machine}  is to create an effective Energy Management Model (EMM) that incorporates renewable energy sources in order to handle the intricate problems associated with energy demand management in prosumer-based smart grids. By modeling complicated correlations inside the stochastic prosumer-based smart grid using Machine Learning (ML), and more specifically Gaussian Process Regression (GPR), the study seeks to simplify EMM. An optimization model for Prosumer Energy Surplus (PES), Prosumer Energy Cost (PEC), and Grid Revenue (GR) is formulated to calculate base performance parameters. Genetic Algorithm (GA) is employed to optimize PES, PEC, and GR under varying scenarios and constraints.

\begin{figure}[H]
\includegraphics[width= 13 cm]{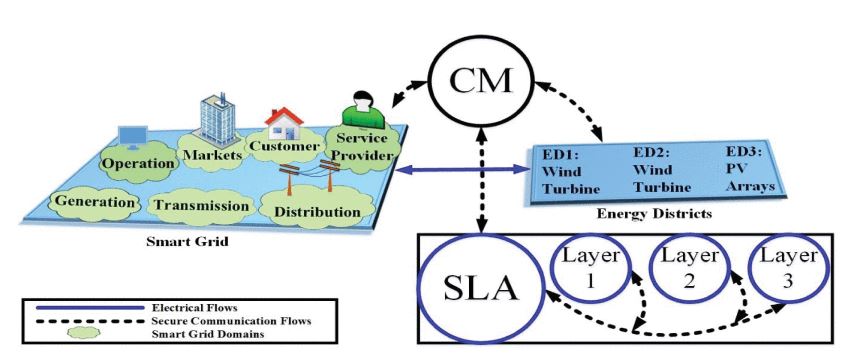}
\caption {Proposed energy management framework for smart grids and EDs. Cluster Manager (CM), Energy District (ED), and Service Level Agreement (SLA) are acronyms \cite{ahmed2020machine}. 
\label{fig10}}
\end{figure}  
\unskip

Figure \ref{fig10} shows the conceptual EMM for bidirectional energy transfer between EDs and the smart grid. The ML-GPR integrated EMM outperforms conventional optimization-based EMM, resulting in reduced energy consumption costs, lower surplus energy, and maximized grid revenue. The adaptive SLA designed in the EMM proves beneficial in achieving improved results, especially in terms of reduced energy costs, surplus energy, and increased grid revenue. 

The authors of the research \cite{karimipour2019deep} aim to address the critical challenges encountered in smart grid control applications, specifically focusing on the intricacies of time synchronization that impede real-time monitoring, measurement, and control. Recognizing the limitations of existing synchronization methods, particularly within IEEE C37.118 and IEC 61850, the authors endeavor to develop an advanced synchronization scheme employing Artificial Intelligence (AI) techniques. The research proposes an innovative AI-based synchronization scheme aimed at mitigating timing challenges encountered in smart grid applications. The introduced scheme leverages a backpropagation neural network as the core AI method, enabling the estimation of precise timing and correction of errors. By incorporating this novel approach, the synchronization scheme seeks to enhance the overall performance and reliability of smart grids, addressing existing issues within the conventional synchronization methods. The utilization of AI techniques, particularly the backpropagation neural network, demonstrates a forward-looking solution to optimize timing accuracy and overcome the complexities associated with real-time communication systems in smart grid environments.

\begin{figure}[H]
\includegraphics[width= 13 cm]{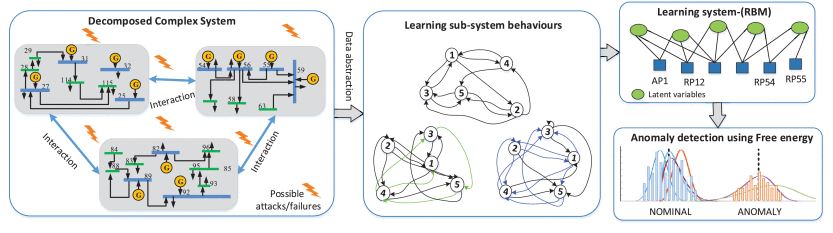}
\caption {Unsupervised learning-based approach for cyberattack detection is proposed. \cite{karimipour2019deep}. 
\label{fig11}}
\end{figure}  
\unskip

Figure \ref{fig11} shows the proposed data-driven architecture for anomaly detection. Initially, the system is divided into several smaller systems. Next, SDF is used to discover the causal relationship between nominal properties of subsystems. The researchers propose an unsupervised anomaly detection approach for smart grids, eliminating the need for labeled data. They introduce a scalable method with reduced computational burden using Spectral Density Function (SDF) and develop a robust learning model based on Deep Belief Networks (DBN) for effective anomaly detection in smart grid systems. This innovative approach holds promise for real-time anomaly detection in dynamic smart grid environments. The suggested algorithm's performance was assessed using several IEEE test systems and operating environments for a number of metrics (TPR, FPR, and ACC). The outcomes showed that the system achieves less than 2\% FPR, 98\% TPR, and 99\% accuracy.

\subsection{Industrial Automation}
\begin{figure}[H]
\includegraphics[width= 8 cm]{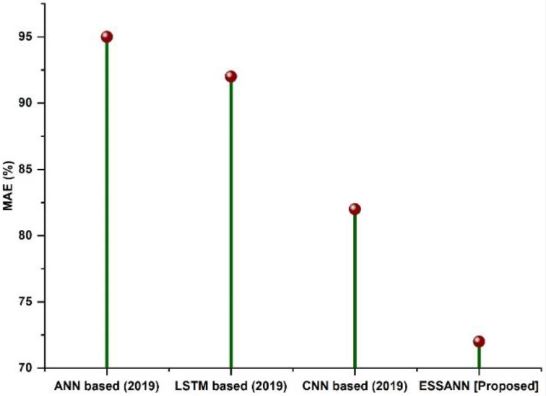}
\caption {Comparing the suggested method's Mean Absolute Error (MAE) with alternative approaches. \cite{al2022machine}. 
\label{industrial}}
\end{figure}

One factor contributing to the latest advancements in Industry 4.0 is artificial intelligence (AI) \cite{javaid2022artificial}. Businesses are concentrating on increasing product uniformity, productivity, and cutting operational costs. They hope to do this through a cooperative relationship between humans and robotics \cite{javaid2022artificial}. Hyperconnected manufacturing processes in smart industries rely on many equipment that communicate with each other through AI automation systems that capture and understand all kinds of data.
The research paper \cite{al2022machine} introduces an innovative approach, the Elaborative Stepwise Stacked Artificial Neural Network (ESSANN) algorithm, aimed at enhancing industrial automation by leveraging advanced technologies. By utilizing an industrial dataset provided by KLEEMANN Greece and employing techniques such as Principal Component Analysis (PCA) for feature extraction and the Least Absolute Shrinkage and Selection Operator (LASSO) for feature selection, the ESSANN algorithm offers significant improvements over traditional methods. The study evaluates the algorithm's performance across various metrics, including delay, network bandwidth, scalability, computation time, packet loss, operational cost, accuracy, precision, recall, and mean absolute error (MAE). Results indicate notable enhancements, including a 52\% reduction in delay, 97\% optimization in network bandwidth, 96\% improvement in scalability, 59-second computation time, minimal packet loss (approximately 53\%), 59\% reduction in operational cost, and impressive accuracy, precision, recall, and MAE values. The findings underscore the effectiveness of the proposed ESSANN algorithm in advancing industrial automation, offering promising prospects for increased efficiency, productivity, and cost-effectiveness across diverse industrial sectors. Fig \ref{industrial} presents a comparison of performance analysis of the suggested method with the current models.
\unskip

\section{AI-Methods Used For Power Consumption} \label{sec3}

The use of AI in the energy industry is not entirely new. Known as "expert systems," the first state-of-the-art evaluation was released in 1989 \cite{zhang1989expert}. In 1997, a number of AI systems utilized in the electrical system were examined \cite{madan1997applications}. In actuality, the energy industry has been using neural networks and rudimentary expert systems for over thirty years now. However, as computing power increased, so did their capabilities, which is why they are now routinely applied to contemporary power system challenges like microgrid operation or monitoring, large-scale power system operation, and forecasting renewable energy \cite{simeunovic2022interpretable}.

By 2050, electricity is predicted to account for more than 50\% of total energy consumption (net zero scenario) \cite{zhakiyev2023optimization}. Therefore, the focus of the current studies is on the applications of AI especially to power systems, given that current trends reveal energy systems evolving into digitalized, electricity-dominated systems \cite{heymann2024reviewing}. AI can support the maintenance of a high degree of confidence in decision-making in the energy industry, which is becoming more and more unexpected, uncertain, complicated, and ambiguous.  Artificial Intelligence (AI) has the potential to facilitate the necessary automation of decision-making in these increasingly complicated market situations \cite{ableitner2020user}. Examples of such activities include revenue allocation at the community energy system level, microgrid load and supply balancing, and unit commitment \cite{lopez2020artificial}. 

Here, we provide few relevant, but not entirety, overview of many studies that discussed the potential of artificial intelligence (AI) and its various techniques to power systems.

\subsection{Fuzzy Logic Control}
The program is possible to use fuzzy logic control (FLC) to control power usage in a variety of systems \cite{ali2023fuzzy}, \cite{dhara2024power}. Because fuzzy logic offers a framework for handling imprecision and uncertainty, it can be used to control systems with complex and nonlinear dynamics, such power consumption, where it may be challenging to establish precise mathematical models.
\begin{figure}[H]
\includegraphics[width= 9 cm]{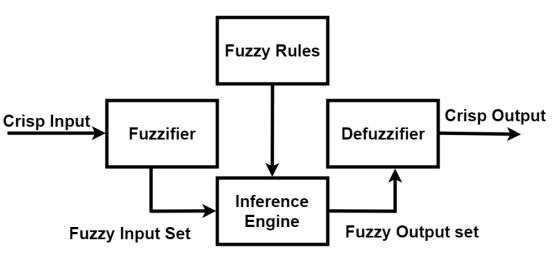}
\caption {Diagram of the fuzzy logic system (FLS).  \cite{shah2020fuzzy}. 
\label{fig12}}
\end{figure}  
\unskip
Fig. \ref{fig12} illustrates a typical fuzzy logic system. When the data is ambiguous, FLS converts linguistics rules based on expert judgment to forecast approximations of the outcomes. FLS can be quickly implemented and doesn't require a mathematical model. The paper \cite{shah2020fuzzy} presented investigates advanced load control strategies aimed at reducing energy consumption and ensuring smooth operation of power systems, with a particular focus on air conditioning loads, which are significant contributors to energy consumption in residential and commercial buildings. Employing fuzzy logic control, the study explores various configurations of membership functions to effectively manage energy consumption while mitigating thermal disturbances. Through extensive simulations conducted in MATLAB Simulink, the fuzzy controller with triangular membership functions emerges as a standout performer, achieving a remarkable 25\% reduction in energy consumption with an error rate of less than 1\%. The suggested plan is illustrated in Fig \ref{fig13}.
\begin{figure}[H]
\includegraphics[width= 11 cm]{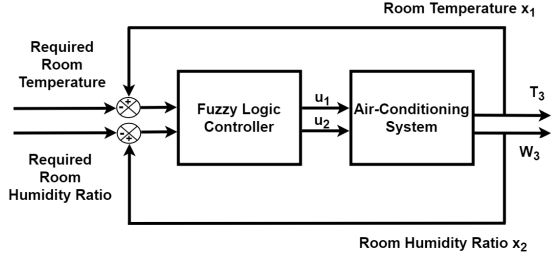}
\caption {Block schematic illustrating the suggested approach. \cite{shah2020fuzzy}. 
\label{fig13}}
\end{figure} 
\unskip
The research paper \cite{boujoudar2023fuzzy} presents an intelligent control strategy for energy management in microgrids, focusing on the effective utilization of lithium-ion batteries and renewable energy sources to stabilize the system and reduce electricity costs. The proposed approach employs a fuzzy logic controller (FLC) to regulate voltage and frequency, ensuring efficient power flow management while considering state-of-charge (SoC) of batteries, solar and wind power generation, and load demand. Through simulations conducted in MATLAB/Simulink, the study demonstrates the capability of the FLC-based energy management system to track desired power levels, regulate DC bus voltage, and maintain microgrid stability under various weather conditions. The bidirectional DC-DC converter plays a central role in maintaining energy balance by controlling battery charging/discharging and power exchanges with the grid. Three operating modes of the battery (charging, discharging, and stop mode) are utilized to optimize energy flow within the microgrid. The proposed fuzzy logic technique proves effective in achieving voltage and frequency regulation while considering multiple system parameters, showcasing its potential as a robust solution for microgrid energy management. Fig. \ref{fig14} is representing the overall methodology of this research. 
\begin{figure}[H]
\includegraphics[width= 13 cm]{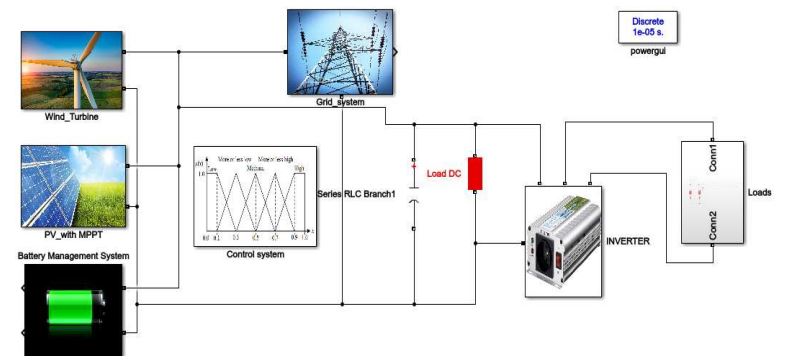}
\caption {The suggested microgrid system \cite{boujoudar2023fuzzy}. 
\label{fig14}}
\end{figure} 
\unskip

\subsection{Reinforcement Learning}
For managing power usage in diverse systems, reinforcement learning (RL) offers a potential solution \cite{perera2021applications}. A machine learning paradigm known as reinforcement learning (RL) teaches an agent to make decisions by interacting with its surroundings in a way that maximizes cumulative rewards. When demand and pricing signals fluctuate, RL algorithms may be used to optimize energy consumption \cite{perera2021applications}. By learning when to charge or discharge based on power costs, grid demand, and the availability of renewable energy, RL may maximize the performance of energy storage devices, such as batteries or capacitors. All things considered, RL provides a flexible and adaptive method of managing power usage, allowing systems to pick up on changes in the surrounding environment, user preferences, and grid requirements. 
\begin{figure}[H]
\includegraphics[width= 7 cm]{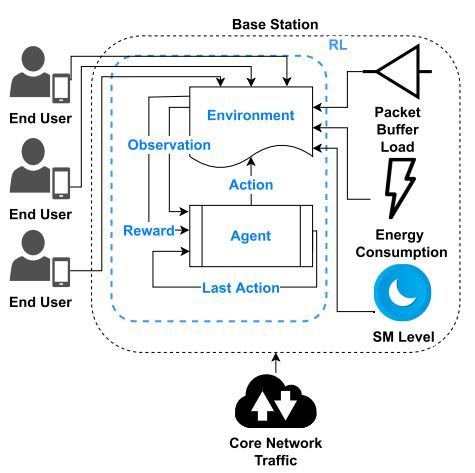}
\caption {The proposed research method's system model \cite{malta2023using}. 
\label{fig15}}
\end{figure} 
\unskip
The research paper \cite{malta2023using} proposes a novel sleep mode management strategy for 5G Base Stations (BSs) aimed at optimizing energy consumption while maintaining Quality of Service (QoS) requirements. The system model (refer to Fig. \ref{fig15}) is characterized in terms of energy consumption, sleep mode regulations, and traffic modulation. Using State Action Reward State Action (SARSA) based reinforcement learning, the approach dynamically adjusts BS components' states to reduce energy consumption during periods of low traffic without compromising latency constraints. Simulation results demonstrate substantial energy savings of up to 80\% in low traffic scenarios, with minimal impact on latency when prioritizing QoS. By considering various traffic loads and latency requirements, the proposed technique ensures end-to-end user latency is met while maximizing energy efficiency. This research offers mobile operators a flexible approach to balancing energy reduction and latency constraints across diverse 5G use cases, contributing significantly to the advancement of sustainable and high-performance mobile networks.

\subsection{Genetic Algorithms}

The authors \cite{wahid2020energy} utilize the Genetic Algorithm (GA) to enhance the optimization process in their proposed approach due to its rich set of operators, including selection, mutation, and crossover, which are well-suited for exploring and exploiting solution spaces effectively \cite{fister2014firefly}. While initially employing the conventional Firefly Algorithm (FA) for energy optimization, the authors find that the solution quality stagnates after a fixed number of iterations, indicating suboptimal results. To address this limitation and further enhance optimization, they integrate the GA into their approach after the termination of the standard FA. This research \cite{wahid2020energy} introduces a novel hybrid model aimed at optimizing energy consumption and enhancing user comfort within residential buildings. By integrating the Firefly Algorithm (FA) and Genetic Algorithm (GA), the proposed approach offers a comprehensive optimization strategy to minimize power usage while maximizing occupant satisfaction. The model considers inputs such as illumination, temperature, air quality, and external environmental factors to fine-tune parameters for optimal comfort and energy efficiency. Utilizing fuzzy controllers, the system dynamically adjusts cooling/heating, illumination, and ventilation systems based on occupant preferences, creating a user-centric environment. Experimental validation on MATLAB R2016a demonstrates the effectiveness of the FA-GA hybrid model compared to conventional optimization techniques. Named FA-GA, the hybrid algorithm iteratively applies FA initially and integrates GA to further enhance optimization. The model addresses multi-objective challenges in energy consumption minimization and Indoor Environmental Quality (IEQ) management within smart buildings, offering a promising solution for sustainable and comfortable living spaces. However, further research is suggested to address limitations such as expanding the consideration of thermal sensation parameters and incorporating additional environmental factors for comprehensive building management systems.
\begin{figure}[H]
\includegraphics[width= 12 cm]{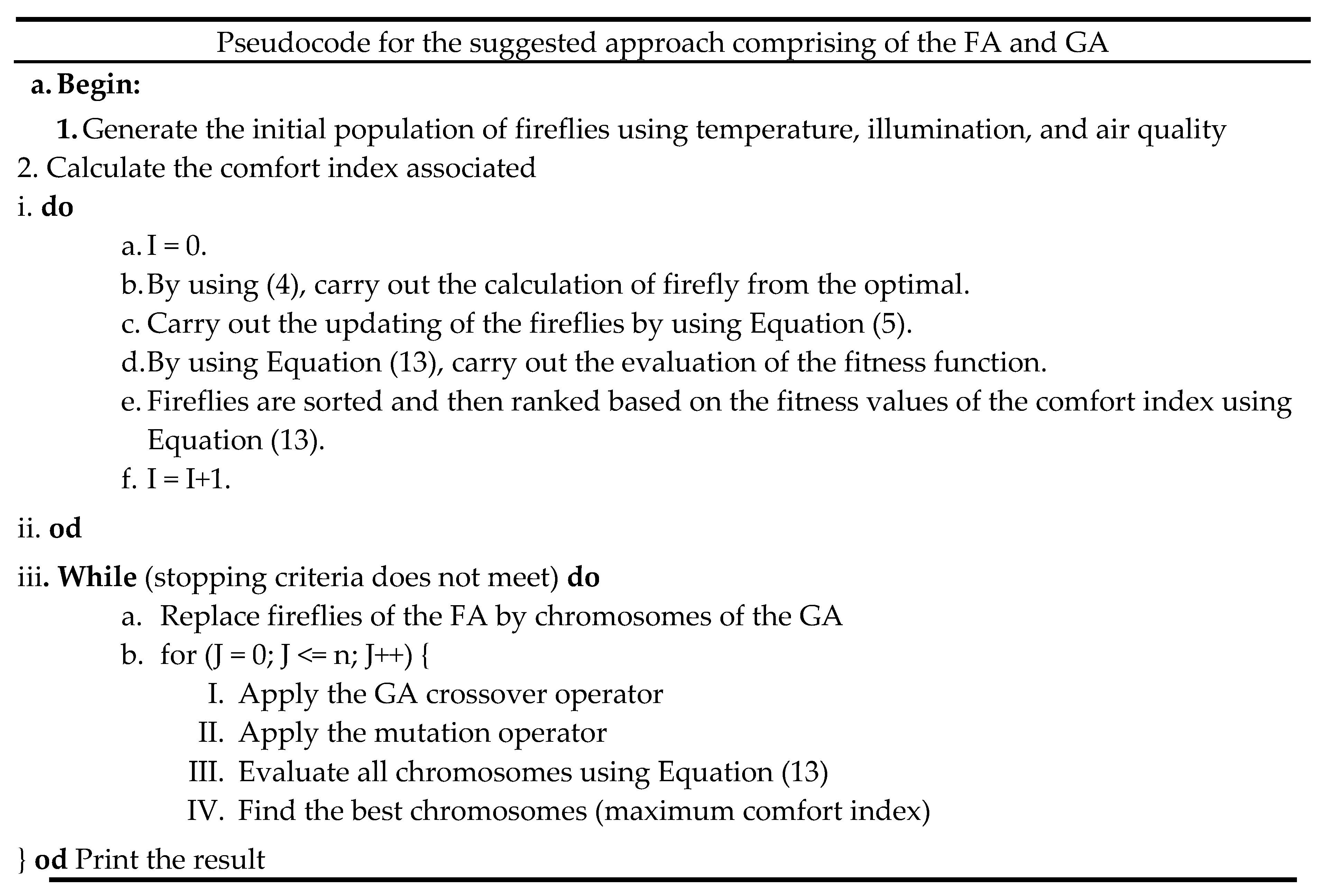}
\caption {The proposed AI approach's pseudocode \cite{wahid2020energy}. 
\label{fig16}}
\end{figure} 
\unskip
Fig \ref{fig16} illustrates the integrated approach utilizing both the Firefly Algorithm (FA) and Genetic Algorithm (GA).
Another research \cite{khan2020genetic} proposes a novel approach for accurate energy consumption forecasting in smart grids, addressing the challenge of developing precise time series models using optimal meteorological features. Leveraging an ensemble machine learning model comprising XGBoost, Support Vector Regressor (SVR), and K-nearest Neighbors (KNN), alongside genetic algorithm (GA) optimization for feature selection, the study achieves superior forecasting performance compared to individual machine learning models. Utilizing electricity consumption data from Jeju Island as a case study, the proposed ensemble model optimized with GA demonstrates enhanced accuracy, resulting in a 3.35\% Mean Absolute Percentage Error (MAPE) for three months of test data. The findings highlight the potential of the proposed model to assist smart grid operators in effectively managing resources and providing superior services to consumers. Key contributions include exploratory data analysis of weather and load consumption data, the introduction of an ensemble machine learning approach with GA-based feature selection, and a comprehensive comparison with different prediction algorithms. This research underscores the significance of optimal feature engineering and machine learning techniques in energy consumption forecasting, offering valuable insights for future improvements and extensions, such as incorporating additional parameters like the number of residents, electric vehicles, and seasonal variations in tourism influx.

\subsection{Swarm Intelligence}
No matter how intelligent an agent is, swarm intelligence (SI) algorithms aim to create positive interactions amongst them \cite{alizadehsani2023swarm}, \cite{abdelaziz2020parallel}. The research \cite{ali2023power} addresses the crucial need for optimal power consumption in healthcare infrastructure, given its increasing dependence on technology such as IoT, Fog Computing, and 5G communications. By introducing a Fog node into an IoT healthcare setup, the study proposes a mathematical formulation to determine the deployment of heterogeneous gateways, aiming to minimize transmission power and infrastructure costs. Utilizing two swarm intelligence-based algorithms, namely the discrete fireworks algorithm (DFWA) and discrete artificial bee colony algorithm (DABC), alongside an ensemble of local search methods, the paper tackles the computationally challenging optimization problem. Comparison with the genetic algorithm (GA) shows promising results, with simulation outcomes demonstrating up to a 33\% reduction in power consumption in the proposed healthcare infrastructure. The study also presents a mathematical model using swarm intelligence-based algorithms to plan a power-aware infrastructure for delay-sensitive healthcare applications, indicating potential extensions to other domains like online gaming, smart cities, and autonomous vehicles. The formula \ref{eq:power_equation} for calculating transmission power is essential as it determines the energy consumption and associated costs in a Power-aware Fog-supported Internet of Healthcare (IoH) network.
\begin{equation}
P_{TX} = P_{RX} \times \left( \frac{4 \pi}{\lambda} \times d \right)^{2}
\label{eq:power_equation}
\end{equation}
The equation \ref{eq:power_equation} represents the received power (\( P_{TX} \)) at a certain distance (\( d \)) from a transmitter, given the transmitted power (\( P_{RX} \)) and the wavelength of the signal (\( \lambda \)).
Overall, this research offers valuable insights into optimizing power consumption in healthcare infrastructure, with implications for improving healthcare data communications and operational efficiency.

\subsection{Machine learning}
Machine learning techniques are increasingly being applied to optimize power consumption \cite{hasan2022new} in various domains \cite{morlans2022power}, including industrial, residential, and commercial sectors. Again, machine learning models can analyze data from smart meters, weather forecasts, occupancy patterns, and building characteristics to optimize heating, cooling, and lighting systems \cite{alzoubi2022machine}. 

The research \cite{tekin2023energy} conducts a comparative analysis of on-device machine learning (ML) algorithms for Intrusion Detection Systems (IDS) in Smart Home Systems (SHSs), focusing on energy consumption for IoT applications. It addresses the security and privacy concerns of cloud-based ML by proposing on-device ML models. The study evaluates training and inference phases separately, comparing cloud, edge, and IoT device-based ML approaches for training, and conventional versus TinyML approaches for inference. Results show that deploying the Decision Tree (DT) algorithm on-device yields better performance in terms of training time, inference time, and power consumption. Energy consumption analyses reveal insights such as the trade-off between accuracy and execution time for different ML algorithms, with Decision Tree and Random Forest exhibiting lower power consumption during training. Additionally, the study highlights the efficiency of cloud computing-based ML for algorithms supporting multicore processing and identifies DT as the optimal choice for inference time performance on IoT devices. The results indicate that k-NN, DT, RF, and ANN outperform LR and NB in terms of accuracy, achieving a high accuracy rate of 99\%. k-NN, DT, RF, and NB also achieve a precision value of 99\%, while LR and ANN achieve 98\%.

\subsection{Neural Networks}
Since neural networks enable the establishment of functional connections between different parameters that are specified as a collection of values for both input and output for the model, they may be utilized to construct models for power consumption forecasting \cite{uakhitova2022electricity}, \cite{mahjoub2022predicting}.

A research article \cite{ruan2022hybrid} introduces a novel approach to energy consumption prediction in the context of building energy management, addressing the challenge of handling complex time series data. By leveraging modal decomposition techniques, specifically variational mode decomposition (VMD), the original energy consumption sequence is decomposed into more robust subsequences, enhancing the predictive accuracy. Additionally, the feature selection process utilizes the maximum relevance minimum redundancy (mRMR) algorithm to analyze the correlation between individual components and features while eliminating redundancy, further improving the model's robustness. The forecasting module employs a long short-term memory (LSTM) \cite{sherstinsky2020fundamentals} neural network model, known for its effectiveness in capturing temporal dependencies in sequential data, to predict power consumption. Comparative experiments conducted on measured data from an office building in Qingdao demonstrate the superiority of the proposed hybrid model. Specifically, the hybrid model exhibits improved prediction accuracy and robustness compared to alternative approaches, resulting in significant performance enhancements. The results reveal that the hybrid model outperforms the VMD-MIM-LSTM approach across various evaluation metrics. The primary procedure of Variational Mode Decomposition (VMD) involves five steps.

Suppose $u_k$ is the $k$th order mode of the original signal $f$, and $\delta(t)$ is a Dirac distribution. The analytic signal of the mode $u_k$ is calculated by the Hilbert transform, then its unilateral frequency spectrum can be expressed as:
    \begin{equation}
        (\delta(t) + j\pi t) \ast u_k(t) 
    \end{equation} \label{eq:equation1}
The resolved mode signal can have its frequency modulated to the matching baseband by appending a pre-calculated center frequency:
\begin{equation}
    \text{Modulated Mode Signal} = \left[ (\delta(t) + j\pi t) \cdot u_k(t) \right] e^{-jwk t}
    \label{eq:modulated_mode_signal}
\end{equation}
Notably, the hybrid model achieves a remarkable improvement in the coefficient of determination (R2) value by up to 10\% and substantial reductions in mean absolute error (MAE) and root mean square error (RMSE) by up to 48.9\% and 54.7\%, respectively, compared to the control groups. These findings underscore the effectiveness of the proposed hybrid model in accurately predicting energy consumption, thereby offering valuable insights for building energy management applications.

\subsection{Predictive Analytics}
The process of obtaining knowledge from data and applying it to forecast patterns and trends is known as predictive analytics \cite{lepenioti2020prescriptive}. In the reseach article \cite{lin2022predictive}, a novel Predictive Power Demand Analytics Methodology (PPDAM) is proposed, leveraging deep neural networks and symbolic aggregate approximation, to forecast power demand patterns in buildings and anticipate both normal (motif) and abnormal (discord) behaviors. The methodology aims to address the challenge of interpreting power demand forecasts by incorporating domain expertise and automated anomaly detection. Experimental results demonstrate that power forecasts can be categorized into distinct demand patterns, each with a specific probability of occurrence. The reliability of anomaly prediction is assessed through a classification test, achieving an accuracy of 88\% and an F1 score of 87.38\%. These findings suggest that the proposed approach offers a promising solution for building operators to extract latent information from power consumption data, potentially enhancing the efficiency and performance of building power systems.

\section{Result Analysis and Future Direction} \label{sec4}
The aim of this review is to examine the contribution of AI algorithms in recent advancements within the field of power research. A comprehensive survey of 17 sources was conducted, focusing on the utilization of diverse AI techniques across various power applications. Additionally, the findings from each survey were compiled into a table, denoted as Table \ref{tab1}, which summarizes the authorship, publication year, methodology, and outcomes of the studies analyzed. This table serves as a valuable resource for tracking the evolution of AI techniques in power research and provides insights into the general trends observed in the surveyed literature.

\begin{longtable}{ @{} p{1.2cm}  p{0.8cm}  p{5cm}  p{5cm}  @{}}
\caption{An overview of this survey article that includes the methodology and results of previous studies}\label{tab1} \\
\toprule
Reference & Year  & Framework & Outcome\\
\midrule
\endfirsthead

\multicolumn{4}{c}%
{{\tablename\ \thetable{} -- continued from previous page}} \\
\toprule
Reference & Year  & Framework & Outcome\\
\midrule
\endhead

\midrule
\multicolumn{4}{r}{{Continued on next page}} \\
\endfoot

\bottomrule
\endlastfoot

\cite{boubaker2023assessment}  & 2023   & Used deep learning (DL) and machine learning (ML) methods for PV module malfunction diagnosis and detection   & An excellent accuracy of 98.71\% was obtained from deep learning approach 
\\
\cite{guo2020fault}  & 2020   &  Proposed fusion model (soft voting and hard voting ensemble model) for distributed photovoltaic  power  station of real-time  fault  diagnosis & The soft voting model provided 94.11\% accuracy  which was proven best among related researchs.  \\

\cite{augbulut2020performance}  & 2020   &  Four machine learning algorithms—support vector machine, artificial neural network, kernel and nearest-neighbor, and deep learning—were used to forecast the results of the photovoltaic investigation. & With extremely satisfactory R2, RMSE, MBE, and MABE values of 0.9921, 0.7082 W, 0.3357 W, and 0.6238 W, respectively, SVM provided superior prediction results.  \\

\cite{kannari2023energy}    & 2023   & Proposed approach to heating regulation that shifts power use and minimizes heating electricity costs is based on reinforcement learning (RL)   &  Using RL method 23\% cost was able to saved. 
\\
\cite{yang2020model}  & 2022   & To regulate the mechanical ventilation and air conditioning systems, a machine learning predictive control system is used. 
& In the two testbeds, the model predictive control saves 36–58\% of the energy.
 \\
\cite{ahmed2020machine}  & 2020   & Suggested combining Gaussian Process Regression (GPR) with machine learning (ML) in the Energy Management Model (EMM)  & The Gaussian Process Regression machine learning algorithm's efficiency is established. A thorough examination of SA-SLAs created in EMM leads to the conclusion that intelligent SLAs with many ROC and ROD associate fewer PEC.
 
  \\
\cite{karimipour2019deep} & 2019   & Developed a scalable anomaly detection engine with Deep Belief Networks (DBN) that is appropriate for large-scale smart grids.  & 99\% accuracy was attained, 98\% of genuine positives were confirmed, and fewer than 2\% of false positives were found. \\
\\
\cite{al2022machine}   & 2022   & Implemented a sophisticated stepwise stacked artificial neural network (ESSANN) technique to significantly enhance the automation sector
& 98.95\% accuracy, 95.02\% precision, 95.02\% recall, and 80\% MAE were attained, along with a latency of around 52\%, 97\% network bandwidth completed, and 59 seconds of computing time. These impressive results were attained. 
 \\
\cite{shah2020fuzzy}  & 2020   & Constructed a fuzzy controller to smooth the energy consumption of an air conditioning load using varying numbers and forms of membership functions  & The triangle membership function results in less than 1\% inaccuracy and 25\% energy savings.
  \\
\cite{boujoudar2023fuzzy}  & 2023   & The suggested method relies on a fuzzy logic controller (FLC) to regulate the microgrid's power and energy management & The outcomes of the simulation demonstrated that the fuzzy logic approach is the most effective means of controlling the microgrid bus's voltage and frequency. \\


\cite{malta2023using}    & 2023   & Proposed a unique sleep mode management technique for 5G Base Stations (BSs) with the goal of maximizing energy use through the application of reinforcement learning  & Achieved 75\% energy savings with 20 ms and 10 ms latencies, and up to 80\% energy savings with 50 ms delay. 
\\
\cite{wahid2020energy}  & 2020   & Suggested the genetic algorithm (GA) and the firefly algorithm (FA) to minimize energy use and increase user comfort in residential structures
& Result showed that 97.43\% comfort index for adding genetic algorithm (GA).
 \\
\cite{khan2020genetic}    & 2020   & Presented a method for accurately estimating power usage using an ensemble machine learning model based on the XGBoost, support vector regressor (SVR), and K-nearest neighbors (KNN) regressor algorithms  & 
By using the suggested model, the authors obtained a 3.35 percent MAPE of the three-month test data.\\

\cite{ali2023power}  & 2023   &Introduced a swarm intelligence-based mathematical algorithm for selecting which of two heterogeneous gateways to install in the healthcare infrastructure
& Simulation results showed that power usage might be reduced by up to 33
 \\
\cite{tekin2023energy}  & 2023   & Intrusion Detection Systems (IDS) that are ML-based are suggested as a solution to security and privacy issues with Smart Home Systems (SHSs) & 
Outperforming to get a high accuracy rate of 99\% include t k-NN, DT, RF, and ANN. While LR and ANN attain 98\% precision, k-NN, DT, RF, and NB also reach 99\%.\\

\cite{ruan2022hybrid}   & 2022   & Proposed a neural network model using long short-term memory (LSTM) to forecast electricity usage  & 
The root mean square error (RMSE) fell by 54.7, 35.5, and 34.1\%, and the mean absolute error (MAE) decreased by 48.9, 41.4, and 35.6\%, respectively.\\

\cite{lin2022predictive}   & 2022   & Developed a unique Predictive Power consumption Analytics Methodology (PPDAM) to forecast future normal (motif) and anomalous (discord) behaviors as well as the pattern profile of power consumption in a building.  & 
The classification test provided 88\% accuracy rate and an 87.38\% F1 score. \\
\end{longtable}

\subsection{Investment opportunities in artificial intelligence}

The energy sector is predicted to see a compound annual growth rate (CAGR) of 22.49\% between 2019 and 2024, with a global use of AI expected to reach \$7.78 billion by that time \cite{ahmad2022energetics}. Without a doubt, looking to capitalize on AI's potential advantages for the energy sector is a popular trend among many of the top energy corporations and investors. According to Business Intelligence and Strategy (BIS) study, by 2024, North America will lead the world in AI energy supply \cite{ahmad2022energetics}. At the same time, Asia-Pacific is expected to have a notable increase in the need for more dispersed power generation. 

Despite the fact that over USD 1.8 trillion was invested globally in power in 2018—the primary area of development remains energy efficiency \cite{iea2017world}. The speed at which energy conservation and renewable energy investments have lagged in recent years suggests that environmental sustainability goals, such as those outlined in the Paris Agreement, have even less room to grow \cite{matemilola2020paris}.

\subsection{Artificial intelligence in energy-saving technology and emerging energy applications}
Power companies are only beginning to use \cite{ahmad2022energetics}. The use of digital technology and applications has increased dramatically in recent years. The influence of digital technologies on industry, way of life, and well-being is growing as they become more compact, efficient, and networked \cite{ahmad2022energetics}. Applications of AI in the energy sector include: power forecasting and managing demand, intelligent energy storage, boosting corporate profitability and lowering power system losses, enhance energy storage management, best use of automation, sensors installed in residential buildings for electric cars, demand response metering and invoicing, the microgrid and local market, makes built-in control possible for local microgrids, physical fault investigation and optimization of the network.

\subsection{Future Challenges and Scopes}
Adopting AI in the smart energy industry faces a variety of bottleneck hurdles, including hazards, lack of data, poor data quality, tuning AI network parameters, inadequate technological infrastructure, shortage of skilled experts, integration difficulties, and legal and regulatory concerns. For building energy systems, failure diagnosis and detection are similarly difficult tasks \cite{ahmad2021artificial}. Various studies acknowledge that among the primary obstacles confronting energy systems are data insecurity and insufficient information \cite{ahmad2021artificial}). System performance and dependability are impacted by subpar sensors, controllers, and controlled devices for data estimation and energy system operation. The energy market presents additional difficulties for the power grid's intricate correlations and strong coupling, as well as for the high data dimensionality and enormous complexity of large-scale simulation grid data \cite{ahmad2021artificial}. Using AI to combine renewable energy sources like wind and concerning grid operations, solar is also complicated and challenging. 

Large data computations are performed more quickly by AI approaches than by numerical optimization techniques with greater processor rates. These characteristics enable AI approaches to further automate power systems and improve their performance. Power system planning, control, and operation are done with the goal of using AI to make their many applications easier. We discussed how AI may be used to address issues with power usage in our survey article. AI algorithms may also be used to tackle issues with system frequency fluctuations, fault rate reduction, voltage profile maintenance to decrease transmission losses, and reactive current minimization in distributed systems to boost power factor and enhance voltage profile.

\section{Conclusion} \label{sec5}
In conclusion, this survey paper provides a comprehensive overview of the integration of artificial intelligence (AI) techniques for power optimization across various application domains. Through an extensive analysis of 17 research methods, including studies on solar system photovoltaics, thermal energy management, smart grids, industrial automation, fuzzy logic control, reinforcement learning, genetic algorithms, swarm intelligence, machine learning, neural networks, and predictive analytics, valuable insights into the strengths and limitations of these approaches have been distilled. The survey outlines the significance of AI-driven algorithms and predictive analytics in enabling real-time monitoring and dynamic modifications to power usage trends, thereby increasing efficiency and sustainability across different sectors. Additionally, future directions in the integration of AI for power consumption optimization are outlined, paving the way for further advancements in this field.

\section*{Declarations}

\subsection* {Funding} A portion of the funding for this research comes from DOEd Grant P116Z220008 (1) and NSF Grant No. 2306109. The author(s) expresses all opinions, findings, and conclusions; the sponsor(s) does not necessarily agree with them.
\subsection*{Conflict of interest} The authors do not declare any conflicts of interest.

\subsection*{Ethics approval} There is no original research involving human subjects, animal subjects, or sensitive data in this review publication. 
\subsection*{Consent to participate} The studies and data included in this review were sourced from publicly available and previously published literature. 
\subsection*{Consent for publication} All authors listed on this review paper have reviewed and approved the final manuscript for submission to Machine Learning. Each author has contributed significantly to the research, writing, and revision of the paper. All authors have read and agreed to the content presented in this review, and their consent for publication is hereby provided.
\subsection* {Availability of data and materials} The data and materials used can be found in the references of this work.
\subsection* {Code availability} The review is based on a comprehensive analysis of publicly available literature, and any references to specific methodologies or algorithms are attributed to the respective original sources. We have strived to provide clear citations and references to enable readers to locate the original works for further examination.
\subsection* {Authors' contributions}
 Conceptualization: [Parag Biswas], [Angona Biswas], [Abdullah Al Nasim]; Methodology: [Abdur Rashid], [Angona Biswas], [Abdullah Al Nasim]; Formal analysis and investigation: [Angona Biswas], [Abdullah Al Nasim]; Writing - original draft preparation: [Angona Biswas][Abdullah Al Nasim], [Abdur Rashid],[Parag Biswas]; Writing - review and editing: [Abdullah Al Nasim], [Abdur Rashid]; Funding acquisition: [Kishor Datta Gupta]; Resources: [Abdur Rashid], [Parag Biswas]; Supervision: [Kishor Datta Gupta]


\bibliography{sn-bibliography}

\end{document}